\documentclass[letterpaper]{article} 
\usepackage[submission]{aaai25}  
\usepackage{times}  
\usepackage{helvet}  
\usepackage{courier}  
\usepackage[hyphens]{url}  
\usepackage{graphicx} 
\usepackage{xcolor}
\usepackage{natbib}  
\usepackage{caption} 
\usepackage{algorithm}
\usepackage{algorithmic}
\usepackage{amsmath}
\usepackage{amssymb}
\usepackage{makecell}
\usepackage{subcaption}

\usepackage{newfloat}
\usepackage{listings}
\DeclareCaptionStyle{ruled}{labelfont=normalfont,labelsep=colon,strut=off} 
\floatstyle{ruled}
\newfloat{listing}{tb}{lst}{}
\floatname{listing}{Listing}
\title{Federated In-Context LLM Agent Learning}
\author{
    Panlong Wu\textsuperscript{\rm 1},
    Kangshuo Li\textsuperscript{\rm 1},
    Junbao Nan\textsuperscript{\rm 1},
    Fangxin Wang\textsuperscript{\rm 1}
}
\affiliations{
    \textsuperscript{\rm 1}The Chinese University of Hong Kong, Shenzhen\\

}

\usepackage{bibentry}

\begin{document}

\maketitle

\begin{abstract}
Large Language Models (LLMs) have revolutionized intelligent services by enabling logical reasoning, tool use, and interaction with external systems as agents. 
The advancement of LLMs is frequently hindered by the scarcity of high-quality data, much of which is inherently sensitive. Federated learning (FL) offers a potential solution by facilitating the collaborative training of distributed LLMs while safeguarding private data. However, FL frameworks face significant bandwidth and computational demands, along with challenges from heterogeneous data distributions. 
The emerging in-context learning capability of LLMs offers a promising approach by aggregating natural language rather than bulky model parameters. Yet, this method risks privacy leakage, as it necessitates the collection and presentation of data samples from various clients during aggregation.
In this paper, we propose a novel privacy-preserving Federated In-Context LLM Agent Learning (FICAL) algorithm, which to our best knowledge for the first work unleashes the power of in-context learning to train diverse LLM agents through FL.   
In our design, knowledge compendiums generated by a novel LLM-enhanced Knowledge Compendiums Generation (KCG) module are transmitted between clients and the server instead of model parameters in previous FL methods. 
Apart from that, an incredible Retrieval Augmented Generation (RAG) based Tool Learning and Utilizing (TLU) module is designed and we incorporate the aggregated global knowledge compendium as a teacher to teach LLM agents the usage of tools. We conducted extensive experiments and the results show that FICAL has competitive performance compared to other SOTA baselines with a significant communication cost decrease of $\mathbf{3.33\times10^5}$ times.

\end{abstract}

\section{Introduction}
The emergence of Large Language Models (LLMs) has introduced a revolutionary approach to address the growing demands for advanced intelligent services. Different from traditional smaller neural networks, LLMs are trained on massive diverse data with billions of parameters thus enabling them to have emergent abilities \cite {wei2022emergent} that traditional neural networks do not have. These emergent abilities enable LLMs to have the ability to carry out logical reasoning and thinking as well as interact with external tools from the open world thus can help them deal with more diverse and complex tasks as agents.

Despite the rapid development of LLMs and the extraordinary abilities they have, the abilities of LLM agents in downstream tasks are often restricted by the amount of high-quality data. However, most data is stored locally and privately thus preventing LLMs from absorbing more data to improve their performance. Federated learning (FL) has emerged as a promising approach, enabling the collaborative improvement of models across multiple clients without the direct exchange of private data. In FL, knowledge sharing among diverse clients is facilitated through the aggregation of model weights thus protecting privacy.

However, training LLM agents in FL leads to significant challenges for real-world deployment. The first challenge is the \textbf{mismatch between the high bandwidths consumption and modern communication system}. Popular LLMs like LLaMA3.1-405B \cite{dubey2024llama} need more than ten hours to transmit between two distributed nodes under A typical 100 Mbps communication network. The LLMs' parameter sharing in FL between clients and the central server in each communication round is a huge burden for modern communication systems. The second challenge is the \textbf{mismatch between high computation consumption and modern computation hardware}. Popular modern LLMs usually have billions of parameters which are thousands of times larger than traditional models while the development of hardware cannot catch up with it. Training LLMs is computationally intensive, resulting in prolonged training times and significant costs due to the necessity of acquiring GPUs with high processing power and large memory capacities. The third challenge is the \textbf{heteogeneious data distribution on different clients}. Different clients may have non-IID data distribution because of the differences in user geographic location or industry. This can harm the aggregation of model parameters \cite{zhao2018federated} and further complicate the FL process.

The in-context learning ability of LLM sheds light on the federated training of LLM agents. With the substantial increase in the parameter size of language models, LLMs demonstrate a remarkable ability to comprehend the provided context and can enhance their performance when additional knowledge is included in the context. This suggests a straightforward approach to enhancing the tool-using capabilities of LLM agents by providing a variety of examples that include instructions, corresponding tools, and input parameters necessary for utilizing these tools within the context, thereby enabling LLM agents to learn from these instances.
The in-context learning capability of LLMs allows them to acquire knowledge extensively from the natural language within the provided context. By integrating this capability into FL, we can transmit natural language context rather than the LLM's cumbersome parameters, thereby significantly reducing communication costs. However, applying in-context learning in FL can lead to the leakage of user privacy as it often requires data samples from different clients to be collected and presented in the context during aggregation.
\textit{Addressing how diverse LLM agents can access knowledge via in-context learning while safeguarding privacy in FL remains an unsolved challenge in prior research.}

In this paper, we propose a novel privacy preservative Federated In-Context LLM Agent Learning (FICAL) algorithm to fill this gap which is to our best knowledge for the first time unleashing the power of in-context learning in FL of LLM agents to address these challenges. \textbf{Different from all previous traditional FL algorithms which transmit model parameters every communication round, in FICAL, we novelly design a Knowledge Compendium Generation (KCG) module to generate knowledge compendiums that contain tool usage knowledge that is transmitted and aggregated.} This novel design enables FICAL to have a communication consumption of $O(1)$ while traditional parameter-sharing FL algorithms have a communication consumption of $O(N)$ with respect to model size. The extraordinary communication performance guarantee also demonstrates that FICAL has significant advantages in scalability as the trend towards larger model parameter sizes continues into the future. Furthermore, we design a Retrieval Augmented Generation (RAG)-based Tool Learning and Utilizing (TLU) module to enable the LLM agent to learn how to use tools through a long-context aggregated knowledge compendium. This TLU module addresses the scalability challenges encountered when FICAL supports a substantial number of clients. Such a situation can lead to an excessively long context in the aggregated knowledge compendium, which may adversely affect the performance of the LLM agent and potentially exceed its maximum context length.

We consider the FL scenario of multiple clients owning local LLM agents and private data of tool-using instances of different tools cooperatively to train a global LLM agent. Our design consists of the following steps. (1) Each client generates a local knowledge compendium through a novelly designed LLM-enhanced Knowledge Compendium Generation (KCG) module based on their local datasets and transmits it to the central server. The local knowledge compendium contains information such as the usage scenario of the tools, precautions for using the tools, coordination of different tools, etc. (2) The central server receives knowledge compendiums collected from different clients, it aggregates them to form a global knowledge compendium and sends them back to clients.
This global compendium contains knowledge that can teach the LLM agent to use tools and is privacy-protective because it is generated to describe the information of tools rather than previous methods that generate synthetic data. These methods often leak the information of private data distribution and are prone to be attacked \cite{slokom2022machine}. (3) Clients receive the global knowledge compendium, they use it as teachers, learn how to use corresponding tools for different queries, and invoke tools through a novel RAG-based Tool Learning and Utilizing (TLU) module.

In summary, the main contributions of this paper can be summarized as follows:
\begin{itemize}
    \item We propose a novel one-round communication-efficient, computation-efficient, and privacy-preservative FL algorithm called Federated In-Context LLM Agent Learning (FICAL), which is to our best knowledge \textbf{the first work to unleash the power of in-context learning in FL of LLM agents}. 
    \item In our design, privacy-preserving local knowledge compendiums which are generated by a novelly designed LLM-enhanced KCG module are transmitted instead of the model parameters in traditional FL. \textbf{FICAL achieves a communication efficiency of $O(1)$ complexity, irrespective of model size, whereas traditional FL incurs a linear $O(N)$ overhead, scaling with the model size.}
    \item We design a Retrieval Augmented Generation (RAG)-based Tool Learning and Utilizing (TLU) module to overcome the long-context issue in knowledge compendium learning and improve the accuracy by $7.6\%$.
    \item We have conducted extensive experiments on different scenarios, and results show that FICAL  can achieve competitive results compared to other SOTA baselines with $\mathbf{3.33\times10^5}$ times communication costs decrease.
    
\end{itemize}

\section{Related Work}

\subsection{Single LLM Agent Learning}
Several works have been done related to LLM agent learning. 
\cite{chen2024reprompt} introduce a novel method to enhance LLM’s ability to plan within specific domains, which employs "gradient descent" to optimize the step-by-step instructions within the prompt of LLM agents, using the chat history from interactions with these agents. \cite{biderman2024emergent} aim to predict which sequences will be memorized before the complete training of a large model by extrapolating the memorization behavior from lower-compute trial runs, enabling us to offer equi-compute recommendations to maximize the reliability of these predictions. \cite{madaan2024self} introduce an approach for enhancing initial outputs from LLMs through iterative feedback and refinement, utilizing a single LLM as the generator, refiner, and feedback provider. 
 
\subsection{Resource efficient LLM Learning}
With the advancing capabilities of LLMs, in-context learning has become a new paradigm in natural language processing. 
\cite{wei-etal-2023-symbol} present a method of fine-tuning language models on in-context input-label pairs where natural language labels (e.g., "positive/negative sentiment") are replaced with arbitrary symbols (e.g., "foo/bar"). This approach boosts performance on unseen in-context learning tasks and provides greater robustness to unspecified prompts. \cite{liu-etal-2022-makes} propose a strategy of selecting in-context learning examples to formulate its corresponding prompt based on similarities between queries and examples where selected examples may serve as more informative inputs intuitively.

Retrieval Augmented Generation (RAG) \cite{lewis2020retrieval} enables LLMs to interact with an external large dataset to enhance their performance by retrieving related knowledge. \cite{asaiself} propose a self-RAG algorithm that utilizes the self-reflection of the LLM to help them improve their performance to avoid unnecessary information in the RAG process. \cite{kim2023tree} propose a Tree of Clarification (TOC) algorithm that can effectively deal with the problem of ambiguity in the open domain by adopting recursive construction of ambiguity resolution trees and self-validation pruning methods. 

\subsection{FL with LLMs} 
Several works have been done related to FL with Large Language Models. \cite{zhang2023fedpetuning} explore the performance and the resource consumption of popular parameter efficient tuning methods such as LoRA \cite{hulora}, adapter \cite{houlsby2019parameter}, and prefix tuning \cite{li2021prefix} under various FL settings.
\cite{sun2024improving} improve the LoRA \cite{hulora} method by fixing the randomly initialized non-zero matrices when differential privacy (DP) is added in FL to guarantee user privacy. \cite{sunfedbpt} reduce the communication cost by training optimal prompts and utilizing gradient-free optimization methods. \cite{peng2024fedpft} propose the sub-FM construction module and the sub-FM alignment module to enhance the performance of FL.

\section{Motivation}

\begin{figure}[t]
   \centering
   \includegraphics[width=0.45\textwidth]{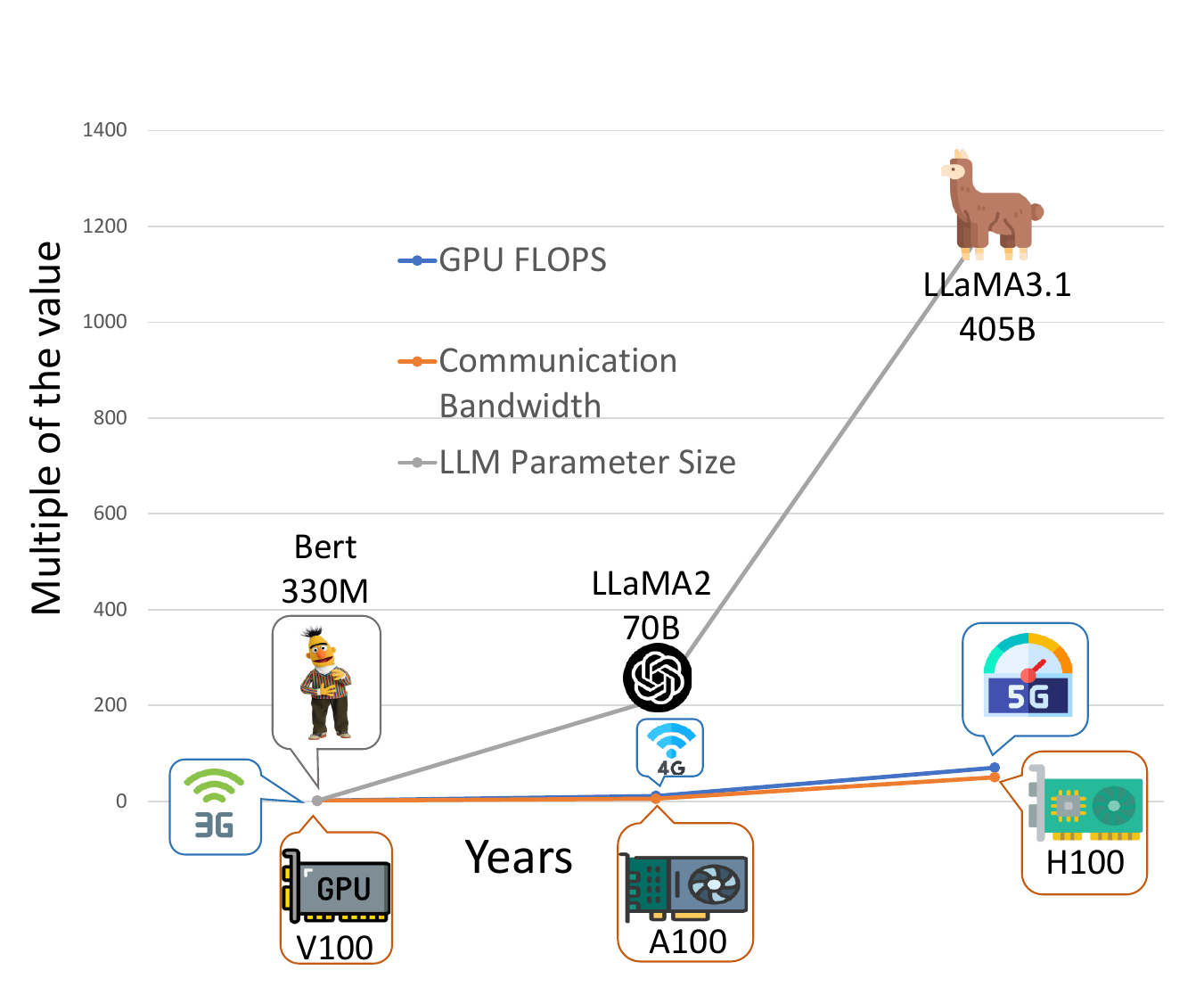}
   \caption{{\textcolor{black}{Development of LLM, communication and computation technology}}}\label{fig:motivation}
\end{figure}

\begin{table}[th]
\centering
\begin{tabular}{|l|c|c|c|c|}
\hline
Model          & LLaMA3.1 & GPT3 & Qwen2 & Mistral-L-2 \\ \hline
Params         & 405 B     & 175 B & 72 B   & 123 B      \\ \hline
\begin{tabular}[c]{@{}l@{}}Trans.\\ Time (h) \end{tabular}  & 36   & 15.56    & 6.4  & 10.93         \\ \hline
\end{tabular}
\caption{\textcolor{black}{Parameter size and transmission time of popular LLMs under modern communication system}}
\label{tab:motivation_1}
\end{table}

Existing LLMs are often very bulky with a parameter size of over one Billion and have a rapidly rising trend. 
In Figure. \ref{fig:motivation}, the grey line shows the parameter size of historically significant LLMs, the blue line illustrates the development of different generations of communication technology and the orange line represents the development of computation devices. From the figure, we can conclude that although all of them show an increasing speed, the rate of increase in the number of parameters surpasses that of communication bandwidth and computation power. This finding suggests that the advancement of communication speed and computational capabilities is unable to keep pace with the growing demands of training LLMs.

Table \ref{tab:motivation_1} shows the parameter size and the transmission time of some popular LLMs under modern communication networks. We select four popular LLMs and take the typical transmission rate of 100 Mbps to calculate the transmission time. From the table, we can observe that it takes 6.4 h to 36 h to transmit these LLMs' parameters across different devices. Additionally, the training of LLMs requires extensive computational power.

Efficient model updates and synchronization across distributed clients rely heavily on powerful computation and rapid communication. However, these requirements are particularly taxing, even for state-of-the-art hardware, which may struggle to manage the high computational load and the extensive data transfer needed to support the FL process. Consequently, the deployment and efficient training of such LLMs in federated settings remain formidable tasks. 

To solve this dilemma, we propose FICAL, whose idea is inspired by the famous Chinese book \textit{Dao De Jing}\footnote{The origin of this proverb is said to have many versions, and here we adopt one of them} written by \textit{Lao-tzu} (a famous philosopher in Chinese history, who is considered the founder of Taoist thought). In this book, there is a saying \textbf{: Give a man a fish and you feed him for a day. Teach him how to fish and you feed him for a lifetime."} The saying underscores the principle that the imparting of knowledge and skills is more significant and enduring than the provision of material aid when helping others. Through education, people can learn how to solve problems. This philosophy inspires us to rethink the federated in-context learning of diverse LLM agents. The most frequently used way of placing data samples in the context can be modified in FL as it can leak the data privacy of clients if they share their data samples. 

In our design, instead of transmitting private data samples, each client transmits a knowledge compendium that encapsulates instructions for utilizing the tools. These compendiums are generated by distilling the essential information required to master the tools from the local datasets. The collection of all local knowledge compendiums forms a global compendium, which comprises comprehensive knowledge of tool usage. This global compendium can then serve as a teacher, instructing LLM agents in tool usage, analogous to the adage of teaching a person how to fish.

\section{Methodology}
\subsection{Problem Statement}
In this paper, we consider a FL scenario with a total number of $N$ clients and a total number of  $M$ types of toolsets, each toolset consists of several tools in a similar domain.
Each client $i$ has its own dataset $D_i$ which consists of tool-using instances. Different clients' datasets may consist of instances of different tools as in practice different participants in FL may only have access to data from certain areas due to constraints of their respective industries (medical, education, sports, etc.).

\subsection{Preliminaries on In-Context Learning}
Different from previous neural networks, LLMs can learn knowledge from the natural language context in the prompt given to them without changing the weights of their parameters \cite{brown2020language}. This novel approach to learning, coined "in-context learning," has garnered significant attention as one of the emergent capabilities of LLMs \cite{wei2022emergent}, setting them apart from traditional neural network architectures.
Consequently, the in-context learning approach results in significant reductions in computational power and GPU memory usage, enhancing the accessibility and efficiency of these models for real-world deployment.

\subsection{Preliminaries on Retrieval Augmented Generation}
Despite the impressive in-context learning capabilities of LLMs, they encounter significant challenges when processing long contexts. As the context length increases, these models may struggle to allocate attention effectively. The Self-Attention mechanism, which relies on focusing on relevant contextual information, can become compromised in long sequences where pertinent information may be diluted or overshadowed by less relevant data, resulting in a dispersion of the LLM's focus.\cite{song2024counting}. 
The RAG process is as follows. When receiving a query, first a retrieval model is used to extract the related information in the vector database with a large amount of data and then pass it to LLMs. Finally, the LLM answer the query equipped with the extracted knowledge from the vector database. This methodology enables LLMs to harness the insights of extensive data without grappling with the complexities associated with long-context processing.

\subsection{Traditional FL}
In traditional FL, clients transmit their model parameters to the central server in every communication round. After all the parameters are received, the server aggregates them to form a global model and sends it back to clients. After several communication rounds, the final global model training will be done.
The learning goal of traditional FL can be expressed as

\begin{equation}
    \min_{\mathbf{w}_i} \mathcal{L} = \frac{1}{N} \sum_{i=1}^N \mathbb{E}_{(\boldsymbol{x}_i, y_i) \sim {d}_i} \mathcal{L}_i(\boldsymbol{x}_i, y_i; \mathbf{w}_i)
\end{equation}
where $\mathcal{L}_i$ represents the loss function of client $i$, $(\boldsymbol{x}_i$ and $y_i$ represents the private data and corresponding answer of client $i$, $\mathbf{w}_i$ denotes the parameters of the LLM of client $i$ and ${d}_i$ denotes the data distribution of the private data of client $i$.

\begin{figure*}[t]
   \centering
   \includegraphics[width=0.95 \textwidth]{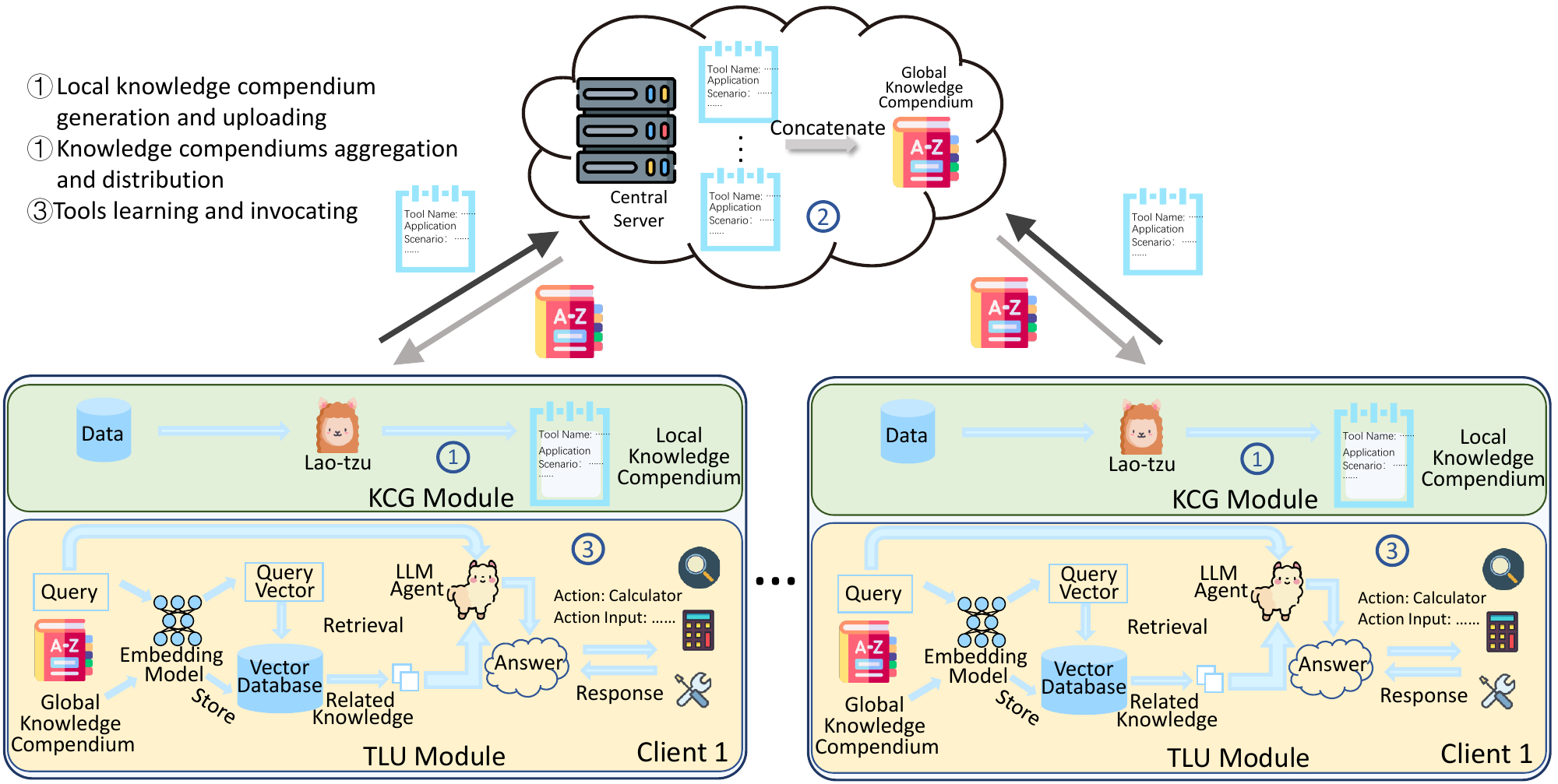}
   \caption{{\textcolor{black}{Workflow of FICAL}}}\label{fig:FICAL}
\end{figure*}

\subsection{FICAL}

The detail of our method is depicted in Figure \ref{fig:FICAL}. The whole process of FICAL can be divided into three parts. 

\textbf{Part one is the generation and uploading of the knowledge compendium.} In this part, each client generates its own local knowledge compendium, denoted as $\zeta_i$ based on its unique local dataset to extract knowledge about how to use the tools. This knowledge extraction procedure is conducted by a novelly designed (Knowledge Compendium Generation) KCG module whose core is an LLM (In this work, we use the DeepSeek-v2 \cite{deepseekai2024deepseekv2strongeconomicalefficient} as the generator) which we called \textit{Lao-tzu} as it generates knowledge compendiums to help LLM agents improve in accordance with the idea proposed by \textit{Lao-tzu}. 

\begin{figure}[h]
   \centering
   \includegraphics[width=0.5\textwidth]{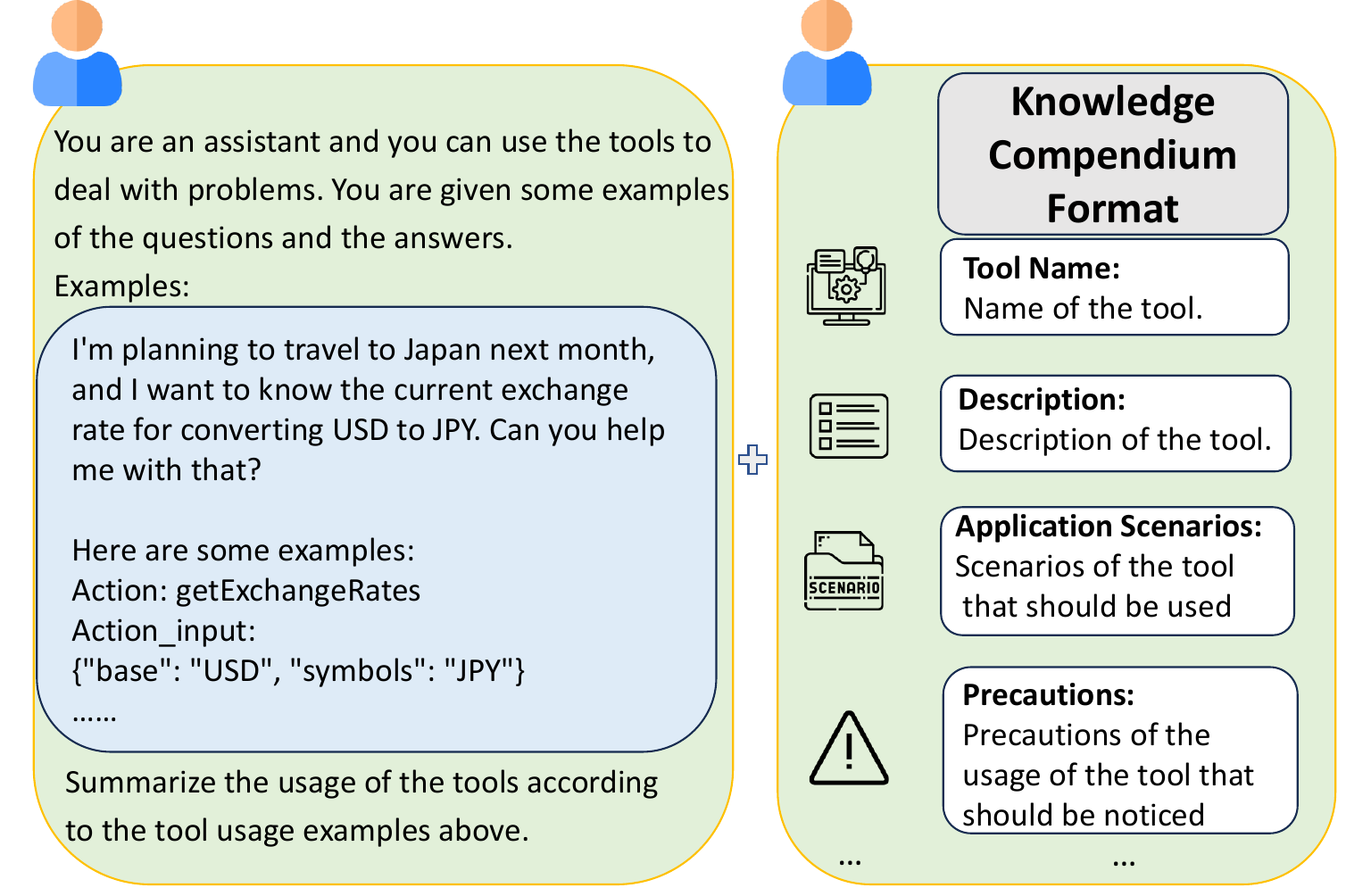}
   \caption{{Knowledge compendium generation template}}\label{fig:template}
\end{figure}

We carefully design a knowledge generation prompt and feed it into \textit{Lao-tzu} to help its' generation. More specifically, Figure \ref{fig:template} shows the design of the knowledge compendium template. In our template, we put the local datasets which consist of instructions from the users, and the corresponding answers that include the name of the tool to be used and input parameters to correctly use the tool as examples.
Given these examples, \textit{Lao-tzu} is going to generate the usage of the tools. This step involves refining knowledge to extract useful data specific to the characteristics of tools from the information observed in examples. The extraction of knowledge about tools offers significant benefits, as the information generated {does not contain private user information (such as personal privacy data)}, yet it aids LLM agents in learning how to use the tools. 
The first part of the generated usage is the \textit{detailed description of the tool}, denoted as $des_i$   which can help users understand the main functions and uses of the tools clearly, so as to determine whether they meet their needs. The second part denoted as $app_i$ is the \textit{application scenarios in which the tools should be used}. This part can help users understand the effectiveness of tools under certain circumstances, and ensure that the appropriate tools are selected to solve specific problems. By understanding the best times and occasions, users can use tools more efficiently to avoid unnecessary attempts. The third part denoted as $pre_i$ is the \textit{precautions of the usage of the tool that should be noticed}. This part guides how to use the API efficiently to avoid wrong requests, thereby improving application performance. Also, it can provide common errors and processing methods to reduce problems that occur when utilizing the tool. The fourth part denoted as $coo_i$, is the \textit{coordination of different tools}. This part can help the tool user to know how to solve the problem through chain calls to tools. The final knowledge compendium can be represented as 
\begin{equation}
    \zeta_i = \{ des_i, app_i, pre_i, coo_i\}
\end{equation}

\begin{figure*}[t]
   \centering
   \begin{minipage}[t]{1\textwidth}
       \centering
       \includegraphics[width=0.9\textwidth]{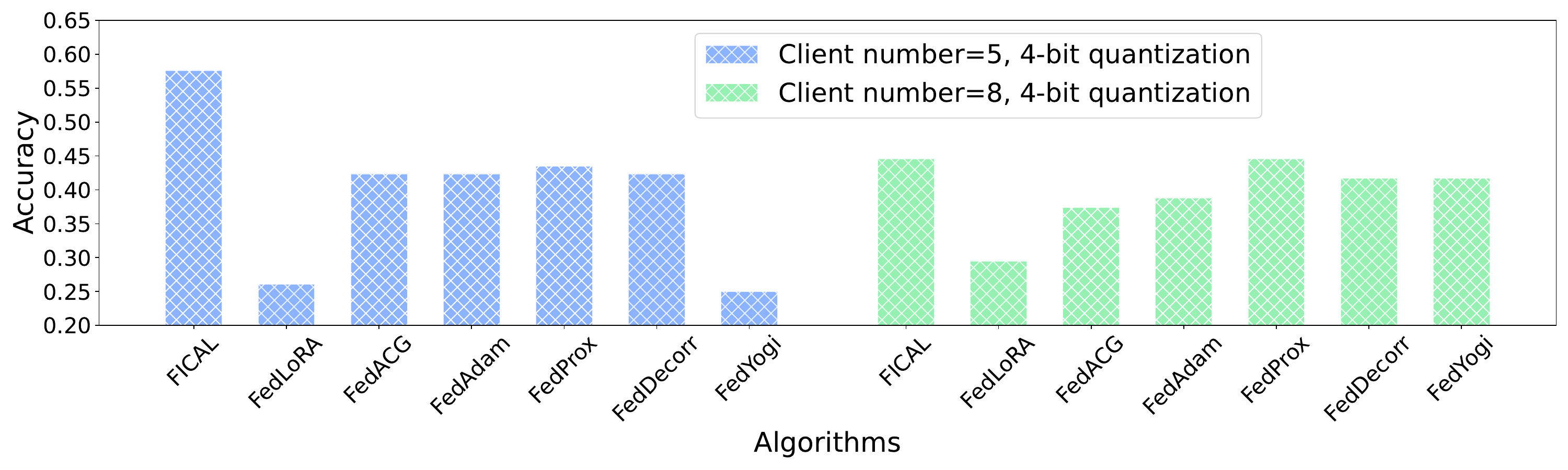}
       \label{fig:4bitclient5_8a}
   \end{minipage}
   \begin{minipage}[t]{1\textwidth}
       \centering
       \includegraphics[width=0.9\textwidth]{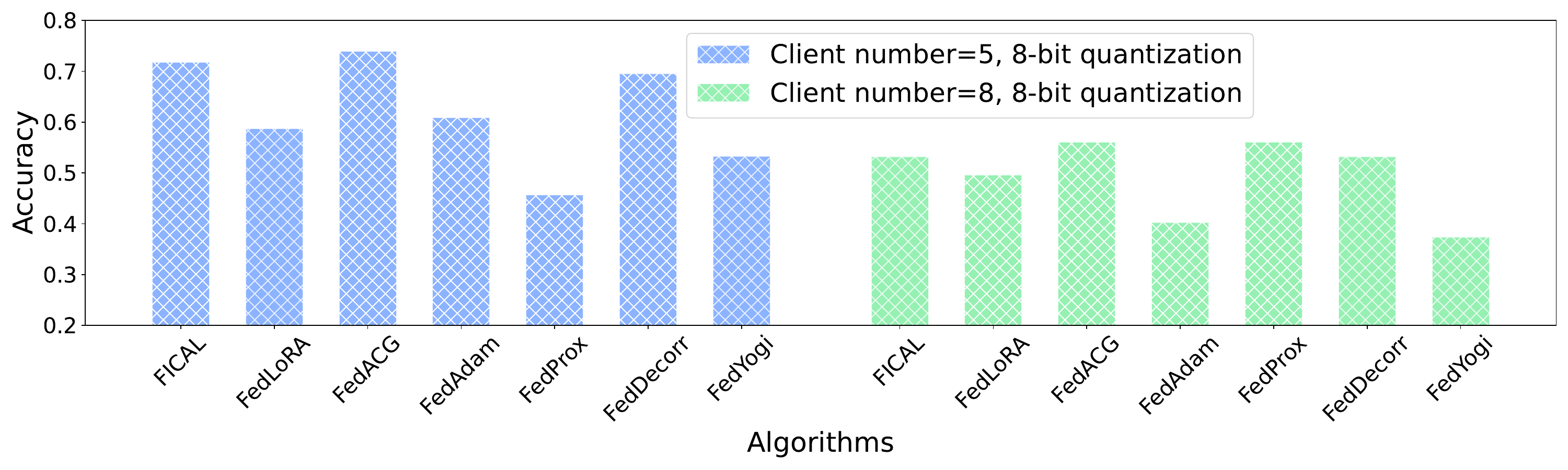}
       \label{fig:8bitclient5_8b}
   \end{minipage}
   \caption{Results comparison under different number of clients }
   \label{fig:combined}
\end{figure*}

\textbf{Part two is the global knowledge compendium generation and offloading.}
After local knowledge compendiums are generated, clients subsequently transmit them to the central server. upon receiving different clients' knowledge compendiums, the central server concatenates them to form a global knowledge compendium $\zeta_g$ which can be represented as 
\begin{equation}
    \zeta_g = \{ \zeta_1, \zeta_2, ..., \zeta_n  \}
\end{equation}
and then sends it back to clients. This global knowledge compendium contains information on how to use diverse tools collected from different clients.

\textbf{Part three is the learning and using of tools}.
When clients collect the global knowledge compendium they first transform them into a vector form using embedding models and then store it on the vector database. In this work, we use the bge-large-en-v1.5 \cite{bge_embedding} as the embedding model. This processed knowledge compendium will act as an external teacher to help LLM agents learn how to use tools. The content of the global knowledge compendium is typically extensive, comprising information gathered from all clients, which poses significant challenges for LLM agents attempting to assimilate this data when directly incorporated into the prompt. This highlights the importance of extracting the necessary information from the global knowledge compendium to answer queries. By extracting relevant information, we can avoid interference from redundant information in the allocation of attention in the attention mechanism, ensuring that useful information for answering queries receives adequate attention. 

We design a novel RAG-based Tool Learning and Utilizing (TLU) module to boost the performance of tool learning. When there is a query to the LLM agent, first the query is converted into a query vector, and then the generated query vector is used to retrieve the most relevant information in the vector database which contains the information on the usage of tools that may be useful to answer the query by calculating the similarity. The extracted related information encompasses knowledge that can instruct the LLM agent in the application of the tools relevant to the query, thereby enhancing its tool learning and utilizing performance.

\begin{table*}[ht]
\centering
\begin{tabular}{|l|c|c|c|c|c|c|c|}
\hline
         & FICAL & FedLoRA & FedACG  & FedAdam & FedProx & FedDecorr & FedYogi \\ \hline
Accuracy & 52.34\%   & 39.25\% & 38.32\% & 35.52\% & 42.99\% & 38.32\%   & 39.25\% \\ \hline
\end{tabular}
\caption{\textcolor{black}{Comparison of tool using accuracy when each client has multiple toolsets' data}}
\label{tab:multitools}
\end{table*}

\begin{table*}[ht]
\centering
\begin{tabular}{|l|c|c|c|c|c|c|c|}
\hline
         & FICAL & FedLoRA & FedACG  & FedAdam & FedProx & FedDecorr & FedYogi \\ \hline
Com. Overhead (MB) & 0.078   & 26000 & 26000 & 26000 & 26000 & 26000   & 26000 \\ \hline
\end{tabular}
\caption{\textcolor{black}{Comparison of communication overhead of different algorithms}}
\label{tab:communication}
\end{table*}

\section{Experiments and Analyses}
In this section, we compare the performance of FICAL to other SOTA baselines under various settings to prove the effectiveness of our algorithm. 
\subsection{Experiments Settings}
\subsubsection{Datasets}
We generate the dataset utilizing the dataset generation method from \cite{tang2023toolalpaca}. We select $M$ tool sets from Public APIs\footnote{https://github.com/public-apis/public-apis}, each tool set consists of a set of tools in a similar domain (for example, a tool set related to dog has tools that can return information of dog types, return how to feed dogs, etc.). The interaction of tools and the LLM agent is in a simulation environment where an external LLM acts as a simulator to simulate the return from the tools. The simulator can get information about the tools and use it as a basis for generating responses. Traces generated from the interaction of the LLM agent and the tools will be fed to a judge LLM (we use the DeepSeek-v2 \cite{deepseekai2024deepseekv2strongeconomicalefficient} in our experiment) to evaluate whether the LLM agent uses the tools correctly.  
The default number of clients is 5, and the default number of toolsets is 5. We use the 4-bit NF4 quantization as the default quantization setting and we assert LoRA parameters to the LLM and only fine-tune LoRA parameters in the baselines to better fit the resource-limited scenario. Extra training on an irrelevant dataset is done on FICAL to improve the LLM agent's ability to follow the output format. All baselines are trained for 50 communication rounds and each client conducts 5 epochs of local training every communication round. All the experiments are done on LLaMA3-8B.

\subsubsection{Baselines}
To verify the effectiveness of the FICAL algorithm,  we compare the successful rate of tool-using with the following state-of-the-art (SOTA) FL baselines.
\textcolor{black}{
\begin{itemize}
    \item \textbf{FedACG \cite{kim2024communication}}: FedACG is a FL algorithm that enhances convergence and stability by broadcasting a global model with a lookahead gradient. 
    \textit{(CVPR 2024)}
    \item \textbf{FedDecorr \cite{shitowards}}: FedDecorr introduces a regularization term during local training that promotes uncorrelated dimensions in the representations.
    \textit{(ICLR 2023)}
    \item \textbf{FedLoRA \cite{peng2024fedpft}}:{FedLoRA is a FL approach that offers a comprehensive empirical study of tuning methods for pre-trained language models in federated environments.} \textit{(ACL 2023 Findings)}
    \item \textbf{FedAdam \cite{reddiadaptive}}: FedAdam is a FL optimization algorithm that adapts the Adam optimizer for distributed training.
    \textit{(ICLR 2021)}
    \item \textbf{FedYogi \cite{reddiadaptive}}:
FedYogi is an FL optimization algorithm that extends the Yogi optimizer to the federated setting. It balances the stability of adaptive gradient methods with the robustness of non-iid data distributions and varying client capabilities, 
\textit{(ICLR 2021)}
    \item \textbf{FedProx \cite{li2020federated}}: FedProx introduces a proximal term in the optimization objective to stabilize training and improve convergence 
    .\textit{(MLSys 2020)}
\end{itemize} 
}

\subsection{Performance Evaluation}

\subsubsection{Comparison of Communication Resource Consumption}
Table \ref{tab:communication} presents a comparison of the communication resource consumption across various algorithms. From the results, we can conclude that FICAL achieves the least communication resource consumption. It can save the communication resource by $\mathbf{3.33\times10^5}$ times compared to other baselines. This is because of the novel knowledge compendium transmission instead of the model parameter sharing in traditional FL which is fatal for models like LLMs which have large parameter sizes. Also, FIACL only requires one round of communication while other baselines always require multiple rounds to converge. Moreover, in traditional FL methods, communication overhead continues to \textbf{increase linearly} with the growth of model parameters, rendering these methods increasingly susceptible to the rising communication cost in the future. In contrast, our algorithm demonstrates that communication overhead \textbf{stays the same} with the expansion of model parameters, underscoring the significant potential of the FICAL algorithm in the future.

\subsubsection{Results when each client has one toolset's data}
We consider the case when each client has one unique toolset and test the accuracy of LLM agent invocating tools. From Figure \ref{fig:combined} we can observe that under the default settings, FICAL achieves the highest accuracy of $57.61\%$ while other baselines achieve accuracies from $26.09\%$ to $43.48\%$ which is $14.13\%$ to $31.52\%$ lower than our algorithm.

\subsubsection{Results when each client has multiple toolsets' data}
We further consider the case that each client has the data of multiple toolsets. More specifically, we consider the case that each client owns two toolsets. Results in Table \ref{tab:multitools} show that FICAL has an accuracy gain of $13.09\%,14.02\%,16.82\%,9.35\%,14.02\%,13.09\%$ compared to other baselines. From these outcomes, we can conclude that FICAL can perform well under different heterogeneous tool data-owning situations which can prove the robustness of our algorithm.

\subsubsection{Results at different quantization level}
We conduct experiments on different quantization levels of the LLM to verify the effectiveness of FICAL. More specifically, we test the performance of different algorithms under 4-bit and 8-bit quantization using the data formats NF4 and NF8. We can conclude from Figure \ref{fig:combined} that When 4-bit quantization is used, FICAL has an accuracy of $57.61\%$ and an accuracy of $44.60\%$ when the number of clients is 5 and 8. Other baselines have an accuracy from $25\%$ to $43.47\%$ and from $24.49\%$ to $44.60\%$. When 8-bit quantization is used, FICAL has an accuracy of $71.74\%$ and $53.24\%$ while other baselines have an 
average accuracy of $60.32\%$ and $48.80\%$. We have discovered that the overall performance under the 8-bit quantization is better than that under the 4-bit quantization, which infers that although quantization can lead to saving in memory, it can cause performance drops due to the loss of precision may cause inaccuracy and even gradient vanishing/explosion in the forward propagation.

\subsubsection{Impact of Number of clients}
To evaluate the performance of the different algorithms under different scales, we test them under different numbers of clients participating in FL. We conduct experiments when there are 5 clients and 8 clients. Results show that FICAL achieves competitive accuracy under different scales.

\begin{table}[]
\begin{tabular}{|c|c|c|}
\hline
         & FICAL(Without RAG) & FICAL With RAG \\ \hline
Accuracy & 50\%               & 57.6\%         \\ \hline
\end{tabular}
\caption{\textcolor{black}{Comparison of accuracies with and without RAG}}
\label{tab:rag}
\end{table}

\subsubsection{Performance comparison with and without RAG}
Table \ref{tab:rag} presents the accuracy of FICAL when employing the RAG-enhanced tool learning module, as well as its performance without the RAG process, which involves directly incorporating the content of the global knowledge compendium into the prompt to instruct LLM agents on tool usage. From the results, we can observe that the FICAL with RAG has an accuracy gain of $7.6\%$ which demonstrates the effectiveness of the TLU module design.

\section{Conclusion}

In this paper, we propose a novel privacy-preserving Federated In-Context LLM Agent Learning (FICAL) which as far as we know the first work to unleash the power of in-context learning in FL of LLM agents. We design a 
LLM-enhanced KCG module to generate privacy-preserving knowledge compendiums on clients and send them to the central server to aggregate into a global knowledge compendium. We additionally design a RAG-based TLU module to enable LLM agents to learn and utilize corresponding tools upon receiving a query. FICAL cut down the communication consumption from $O(N)$ with respect to model size in previous FL methods to $O(1)$, which demonstrates its' tremendous potential in future LLM FL research. We have conducted extensive experiments and results show that FICAL has competitive performance with a communication cost decrease of $\mathbf{3.33\times10^5}$ times.

\bigskip

\bibliography{aaai25}

\begin{thebibliography}{28}
\providecommand{\natexlab}[1]{#1}

\bibitem[{Asai et~al.(2024)Asai, Wu, Wang, Sil, and Hajishirzi}]{asaiself}
Asai, A.; Wu, Z.; Wang, Y.; Sil, A.; and Hajishirzi, H. 2024.
\newblock Self-RAG: Learning to Retrieve, Generate, and Critique through Self-Reflection.
\newblock In \emph{The Twelfth International Conference on Learning Representations}.

\bibitem[{Biderman et~al.(2024)Biderman, Prashanth, Sutawika, Schoelkopf, Anthony, Purohit, and Raff}]{biderman2024emergent}
Biderman, S.; Prashanth, U.; Sutawika, L.; Schoelkopf, H.; Anthony, Q.; Purohit, S.; and Raff, E. 2024.
\newblock Emergent and predictable memorization in large language models.
\newblock \emph{Advances in Neural Information Processing Systems}, 36.

\bibitem[{Brown et~al.(2020)Brown, Mann, Ryder, Subbiah, Kaplan, Dhariwal, Neelakantan, Shyam, Sastry, Askell et~al.}]{brown2020language}
Brown, T.; Mann, B.; Ryder, N.; Subbiah, M.; Kaplan, J.~D.; Dhariwal, P.; Neelakantan, A.; Shyam, P.; Sastry, G.; Askell, A.; et~al. 2020.
\newblock Language models are few-shot learners.
\newblock \emph{Advances in neural information processing systems}, 33: 1877--1901.

\bibitem[{Chen, Koenig, and Dilkina(2024)}]{chen2024reprompt}
Chen, W.; Koenig, S.; and Dilkina, B. 2024.
\newblock RePrompt: Planning by Automatic Prompt Engineering for Large Language Models Agents.
\newblock \emph{arXiv preprint arXiv:2406.11132}.

\bibitem[{DeepSeek-AI et~al.(2024)DeepSeek-AI, Liu, Feng, Wang, Wang, Liu, Zhao, Dengr, and et~al}]{deepseekai2024deepseekv2strongeconomicalefficient}
DeepSeek-AI; Liu, A.; Feng, B.; Wang, B.; Wang, B.; Liu, B.; Zhao, C.; Dengr, C.; and et~al, C.~R. 2024.
\newblock DeepSeek-V2: A Strong, Economical, and Efficient Mixture-of-Experts Language Model.
\newblock arXiv:2405.04434.

\bibitem[{Dubey et~al.(2024)Dubey, Jauhri, Pandey, Kadian, Al-Dahle, Letman, Mathur, Schelten, Yang, Fan et~al.}]{dubey2024llama}
Dubey, A.; Jauhri, A.; Pandey, A.; Kadian, A.; Al-Dahle, A.; Letman, A.; Mathur, A.; Schelten, A.; Yang, A.; Fan, A.; et~al. 2024.
\newblock The llama 3 herd of models.
\newblock \emph{arXiv preprint arXiv:2407.21783}.

\bibitem[{Houlsby et~al.(2019)Houlsby, Giurgiu, Jastrzebski, Morrone, De~Laroussilhe, Gesmundo, Attariyan, and Gelly}]{houlsby2019parameter}
Houlsby, N.; Giurgiu, A.; Jastrzebski, S.; Morrone, B.; De~Laroussilhe, Q.; Gesmundo, A.; Attariyan, M.; and Gelly, S. 2019.
\newblock Parameter-efficient transfer learning for NLP.
\newblock In \emph{International conference on machine learning}, 2790--2799. PMLR.

\bibitem[{Hu et~al.(2022)Hu, Wallis, Allen-Zhu, Li, Wang, Wang, Chen et~al.}]{hulora}
Hu, E.~J.; Wallis, P.; Allen-Zhu, Z.; Li, Y.; Wang, S.; Wang, L.; Chen, W.; et~al. 2022.
\newblock LoRA: Low-Rank Adaptation of Large Language Models.
\newblock In \emph{International Conference on Learning Representations}.

\bibitem[{Kim, Kim, and Han(2024)}]{kim2024communication}
Kim, G.; Kim, J.; and Han, B. 2024.
\newblock Communication-efficient federated learning with accelerated client gradient.
\newblock In \emph{Proceedings of the IEEE/CVF Conference on Computer Vision and Pattern Recognition}, 12385--12394.

\bibitem[{Kim et~al.(2023)Kim, Kim, Jeon, Park, and Kang}]{kim2023tree}
Kim, G.; Kim, S.; Jeon, B.; Park, J.; and Kang, J. 2023.
\newblock Tree of clarifications: Answering ambiguous questions with retrieval-augmented large language models.
\newblock \emph{arXiv preprint arXiv:2310.14696}.

\bibitem[{Lewis et~al.(2020)Lewis, Perez, Piktus, Petroni, Karpukhin, Goyal, K{\"u}ttler, Lewis, Yih, Rockt{\"a}schel et~al.}]{lewis2020retrieval}
Lewis, P.; Perez, E.; Piktus, A.; Petroni, F.; Karpukhin, V.; Goyal, N.; K{\"u}ttler, H.; Lewis, M.; Yih, W.-t.; Rockt{\"a}schel, T.; et~al. 2020.
\newblock Retrieval-augmented generation for knowledge-intensive nlp tasks.
\newblock \emph{Advances in Neural Information Processing Systems}, 33: 9459--9474.

\bibitem[{Li et~al.(2020)Li, Sahu, Zaheer, Sanjabi, Talwalkar, and Smith}]{li2020federated}
Li, T.; Sahu, A.~K.; Zaheer, M.; Sanjabi, M.; Talwalkar, A.; and Smith, V. 2020.
\newblock Federated optimization in heterogeneous networks.
\newblock \emph{Proceedings of Machine learning and systems}, 2: 429--450.

\bibitem[{Li and Liang(2021)}]{li2021prefix}
Li, X.~L.; and Liang, P. 2021.
\newblock Prefix-Tuning: Optimizing Continuous Prompts for Generation.
\newblock In \emph{Proceedings of the 59th Annual Meeting of the Association for Computational Linguistics and the 11th International Joint Conference on Natural Language Processing (Volume 1: Long Papers)}, 4582--4597.

\bibitem[{Liu et~al.(2022)Liu, Shen, Zhang, Dolan, Carin, and Chen}]{liu-etal-2022-makes}
Liu, J.; Shen, D.; Zhang, Y.; Dolan, B.; Carin, L.; and Chen, W. 2022.
\newblock What Makes Good In-Context Examples for {GPT}-3?
\newblock In Agirre, E.; Apidianaki, M.; and Vuli{\'c}, I., eds., \emph{Proceedings of Deep Learning Inside Out (DeeLIO 2022): The 3rd Workshop on Knowledge Extraction and Integration for Deep Learning Architectures}, 100--114. Dublin, Ireland and Online: Association for Computational Linguistics.

\bibitem[{Madaan et~al.(2024)Madaan, Tandon, Gupta, Hallinan, Gao, Wiegreffe, Alon, Dziri, Prabhumoye, Yang et~al.}]{madaan2024self}
Madaan, A.; Tandon, N.; Gupta, P.; Hallinan, S.; Gao, L.; Wiegreffe, S.; Alon, U.; Dziri, N.; Prabhumoye, S.; Yang, Y.; et~al. 2024.
\newblock Self-refine: Iterative refinement with self-feedback.
\newblock \emph{Advances in Neural Information Processing Systems}, 36.

\bibitem[{Peng et~al.(2024)Peng, Fan, Chen, Wang, Pan, Wen, Zhang, and Wang}]{peng2024fedpft}
Peng, Z.; Fan, X.; Chen, Y.; Wang, Z.; Pan, S.; Wen, C.; Zhang, R.; and Wang, C. 2024.
\newblock FedPFT: Federated Proxy Fine-Tuning of Foundation Models.
\newblock \emph{arXiv preprint arXiv:2404.11536}.

\bibitem[{Reddi et~al.(2021)Reddi, Charles, Zaheer, Garrett, Rush, Kone{\v{c}}n{\`y}, Kumar, and McMahan}]{reddiadaptive}
Reddi, S.~J.; Charles, Z.; Zaheer, M.; Garrett, Z.; Rush, K.; Kone{\v{c}}n{\`y}, J.; Kumar, S.; and McMahan, H.~B. 2021.
\newblock Adaptive Federated Optimization.
\newblock In \emph{International Conference on Learning Representations}.

\bibitem[{Shi et~al.(2023)Shi, Liang, Zhang, Tan, and Bai}]{shitowards}
Shi, Y.; Liang, J.; Zhang, W.; Tan, V.; and Bai, S. 2023.
\newblock Towards Understanding and Mitigating Dimensional Collapse in Heterogeneous Federated Learning.
\newblock In \emph{The Eleventh International Conference on Learning Representations}.

\bibitem[{Slokom, de~Wolf, and Larson(2022)}]{slokom2022machine}
Slokom, M.; de~Wolf, P.-P.; and Larson, M. 2022.
\newblock When machine learning models leak: an exploration of synthetic training data.
\newblock In \emph{International Conference on Privacy in Statistical Databases}, 283--296. Springer.

\bibitem[{Song, Zheng, and Luo(2024)}]{song2024counting}
Song, M.; Zheng, M.; and Luo, X. 2024.
\newblock Counting-stars: A simple, efficient, and reasonable strategy for evaluating long-context large language models.
\newblock \emph{arXiv preprint arXiv:2403.11802}.

\bibitem[{Sun et~al.(2024{\natexlab{a}})Sun, Xu, Yin, Yang, Xu, Liu, Du, Chen, and Roth}]{sunfedbpt}
Sun, J.; Xu, Z.; Yin, H.; Yang, D.; Xu, D.; Liu, Y.; Du, Z.; Chen, Y.; and Roth, H.~R. 2024{\natexlab{a}}.
\newblock FedBPT: Efficient Federated Black-box Prompt Tuning for Large Language Models.
\newblock In \emph{Forty-first International Conference on Machine Learning}.

\bibitem[{Sun et~al.(2024{\natexlab{b}})Sun, Li, Li, and Ding}]{sun2024improving}
Sun, Y.; Li, Z.; Li, Y.; and Ding, B. 2024{\natexlab{b}}.
\newblock Improving loRA in privacy-preserving federated learning.
\newblock \emph{arXiv preprint arXiv:2403.12313}.

\bibitem[{Tang et~al.(2023)Tang, Deng, Lin, Han, Liang, Cao, and Sun}]{tang2023toolalpaca}
Tang, Q.; Deng, Z.; Lin, H.; Han, X.; Liang, Q.; Cao, B.; and Sun, L. 2023.
\newblock Toolalpaca: Generalized tool learning for language models with 3000 simulated cases.
\newblock \emph{arXiv preprint arXiv:2306.05301}.

\bibitem[{Wei et~al.(2023)Wei, Hou, Lampinen, Chen, Huang, Tay, Chen, Lu, Zhou, Ma, and Le}]{wei-etal-2023-symbol}
Wei, J.; Hou, L.; Lampinen, A.; Chen, X.; Huang, D.; Tay, Y.; Chen, X.; Lu, Y.; Zhou, D.; Ma, T.; and Le, Q. 2023.
\newblock Symbol tuning improves in-context learning in language models.
\newblock In Bouamor, H.; Pino, J.; and Bali, K., eds., \emph{Proceedings of the 2023 Conference on Empirical Methods in Natural Language Processing}, 968--979. Singapore: Association for Computational Linguistics.

\bibitem[{Wei et~al.(2022)Wei, Tay, Bommasani, Raffel, Zoph, Borgeaud, Yogatama, Bosma, Zhou, Metzler et~al.}]{wei2022emergent}
Wei, J.; Tay, Y.; Bommasani, R.; Raffel, C.; Zoph, B.; Borgeaud, S.; Yogatama, D.; Bosma, M.; Zhou, D.; Metzler, D.; et~al. 2022.
\newblock Emergent Abilities of Large Language Models.
\newblock \emph{Transactions on Machine Learning Research}.

\bibitem[{Xiao et~al.(2023)Xiao, Liu, Zhang, and Muennighoff}]{bge_embedding}
Xiao, S.; Liu, Z.; Zhang, P.; and Muennighoff, N. 2023.
\newblock C-Pack: Packaged Resources To Advance General Chinese Embedding.
\newblock arXiv:2309.07597.

\bibitem[{Zhang et~al.(2023)Zhang, Yang, Dai, Wang, Yu, Qu, and Xu}]{zhang2023fedpetuning}
Zhang, Z.; Yang, Y.; Dai, Y.; Wang, Q.; Yu, Y.; Qu, L.; and Xu, Z. 2023.
\newblock Fedpetuning: When federated learning meets the parameter-efficient tuning methods of pre-trained language models.
\newblock In \emph{Annual Meeting of the Association of Computational Linguistics 2023}, 9963--9977. Association for Computational Linguistics (ACL).

\bibitem[{Zhao et~al.(2018)Zhao, Li, Lai, Suda, Civin, and Chandra}]{zhao2018federated}
Zhao, Y.; Li, M.; Lai, L.; Suda, N.; Civin, D.; and Chandra, V. 2018.
\newblock Federated learning with non-iid data.
\newblock \emph{arXiv preprint arXiv:1806.00582}.

\end{thebibliography}

\pagestyle{empty}
\section*{AAAI Reproducibility Checklist}

Unless specified otherwise, please answer Yes to each question if the relevant information is described either in the paper itself or in a technical appendix with an explicit reference from the main paper. If you wish to explain an answer further, please do so in a section titled “Reproducibility Checklist” at the end of the technical appendix.

This paper:
\begin{itemize}
    \item Includes a conceptual outline and/or pseudocode description of AI methods introduced(Yes)
    \item Clearly delineates statements that are opinions, hypothesis, and speculation from objective facts and results (Yes)
    \item Provides well-marked pedagogical references for less-familiare readers to gain background necessary to replicate the paper (Yes)
\end{itemize}

Does this paper make theoretical contributions? (Yes)

If yes, please complete the list below.
\begin{itemize}
    \item All assumptions and restrictions are stated clearly and formally. (Yes)
    \item All novel claims are stated formally (e.g., in theorem statements). (Yes)
    \item Proofs of all novel claims are included. (Yes)
    \item Proof sketches or intuitions are given for complex and/or novel results. (Yes)
    \item Appropriate citations to theoretical tools used are given. (Yes)
    \item All theoretical claims are demonstrated empirically to hold. (Yes)
    \item All experimental code used to eliminate or disprove claims is included. (Yes)
\end{itemize}

Does this paper rely on one or more datasets? (Yes)

If yes, please complete the list below.
\begin{itemize}
    \item A motivation is given for why the experiments are conducted on the selected datasets (Yes)
    \item All novel datasets introduced in this paper are included in a data appendix. (No)
    \item All novel datasets introduced in this paper will be made publicly available upon publication of the paper with a license that allows free usage for research purposes. (Yes)
    \item All datasets drawn from the existing literature (potentially including authors’ own previously published work) are accompanied by appropriate citations. (Yes)
    \item All datasets drawn from the existing literature (potentially including authors’ own previously published work) are publicly available. (Yes)
    \item All datasets that are not publicly available are described in detail, with explanation why publicly available alternatives are not scientifically satisficing. (NA)
\end{itemize}
Does this paper include computational experiments? (Yes)

If yes, please complete the list below.
\begin{itemize}
    \item Any code required for pre-processing data is included in the appendix.(No).
    \item All source code required for conducting and analyzing the experiments is included in a code appendix. (No)
    \item All source code required for conducting and analyzing the experiments will be made publicly available upon publication of the paper with a license that allows free usage for research purposes. (Yes)
    \item All source code implementing new methods have comments detailing the implementation, with references to the paper where each step comes from (Yes)
    \item If an algorithm depends on randomness, then the method used for setting seeds is described in a way sufficient to allow replication of results. (Yes)
    \item This paper specifies the computing infrastructure used for running experiments (hardware and software), including GPU/CPU models; amount of memory; operating system; names and versions of relevant software libraries and frameworks. (Yes)
    \item This paper formally describes evaluation metrics used and explains the motivation for choosing these metrics. (Yes)
    \item This paper states the number of algorithm runs used to compute each reported result. (Yes)
    \item Analysis of experiments goes beyond single-dimensional summaries of performance (e.g., average; median) to include measures of variation, confidence, or other distributional information. (Yes)
    \item The significance of any improvement or decrease in performance is judged using appropriate statistical tests (e.g., Wilcoxon signed-rank). (Yes)
    \item This paper lists all final (hyper-)parameters used for each model/algorithm in the paper’s experiments. (Yes)
    \item This paper states the number and range of values tried per (hyper-) parameter during development of the paper, along with the criterion used for selecting the final parameter setting. (Yes)
\end{itemize}

\end{document}